%% file: sample.tex
\documentclass[sigconf]{aamas} 

\usepackage{balance} 

\usepackage{algorithm}
\usepackage{algpseudocode}
\usepackage{amsmath}
\usepackage{adjustbox}
\usepackage[bottom]{footmisc}

\setcopyright{ifaamas}
\acmConference[AAMAS '23]{Proc.\@ of the 22nd International Conference
on Autonomous Agents and Multiagent Systems (AAMAS 2023)}{May 29 -- June 2, 2023}
{London, United Kingdom}{A.~Ricci, W.~Yeoh, N.~Agmon, B.~An (eds.)}
\copyrightyear{2023}
\acmYear{2023}
\acmDOI{}
\acmPrice{}
\acmISBN{}

\acmSubmissionID{491}

\title[]{Safe Deep Reinforcement Learning\\by Verifying Task-Level Properties}

\author{Enrico Marchesini*}
\affiliation{
  \institution{Northeastern University}
  \city{Boston (MA)}
  \country{USA}}

\email{e.marchesini@northeastern.edu}

\author{Luca Marzari*}
\affiliation{
  \institution{University of Verona}
  \city{Verona}
  \country{Italy}}
\email{luca.marzari@univr.it}

\author{Alessandro Farinelli}
\affiliation{
  \institution{University of Verona}
  \city{Verona}
  \country{Italy}}
\email{alessandro.farinelli@univr.it}
\author{Christopher Amato}
\affiliation{
  \institution{Northeastern University}
  \city{Boston (MA)}
  \country{USA}}
\email{c.amato@northeastern.edu}

\input{Sections/abstract.tex}


\newcommand{\BibTeX}{\rm B\kern-.05em{\sc i\kern-.025em b}\kern-.08em\TeX}

\begin{document}


\pagestyle{fancy}
\fancyhead{}


\maketitle 


\input{Sections/introduction.tex}

\input{Sections/preliminaries.tex}
\input{Sections/methods.tex}

\input{Sections/experiments.tex}
\input{Sections/relatedwork.tex}
\input{Sections/discussion.tex}


\newpage
\section{Acknowledgements}
This work was partially funded by the Army Research Office under award number W911NF20-1-0265.
\bibliographystyle{ACM-Reference-Format} 
\bibliography{sample}



\end{document}

%% file: Sections/abstract.tex
\begin{abstract}
Cost functions are commonly employed in Safe Deep Reinforcement Learning (DRL). However, the cost is typically encoded as an indicator function due to the difficulty of quantifying the risk of policy decisions in the state space. Such an encoding requires the agent to visit numerous unsafe states to learn a cost-value function to drive the learning process toward safety. Hence, increasing the number of unsafe interactions and decreasing sample efficiency. In this paper, we investigate an alternative approach that uses domain knowledge to quantify the risk in the proximity of such states by defining a \textit{violation} metric. This metric is computed by verifying task-level properties, shaped as input-output conditions, and it is used as a penalty to bias the policy away from unsafe states without learning an additional value function. We investigate the benefits of using the violation metric in standard Safe DRL benchmarks and robotic mapless navigation tasks. The navigation experiments bridge the gap between Safe DRL and robotics, introducing a framework that allows rapid testing on real robots. Our experiments show that policies trained with the violation penalty achieve higher performance over Safe DRL baselines and significantly reduce the number of visited unsafe states.
\end{abstract}

%% file: Sections/introduction.tex
\section{Introduction}

Safe Deep Reinforcement Learning (DRL) approaches typically foster safety by limiting the accumulation of costs caused by unsafe interactions \cite{Garcia2015}. Defining informative cost functions, however, has the same issues as designing rewards \cite{DiversityExpl} due to the difficulty of quantifying the risk around unsafe states. For this reason, recent works rely on indicator functions, where a positive value deems a state unsafe \cite{ConstrainedDRL}. In detail, the cost refers to a state-action pair $(s, a)$, and it is backed up to propagate safety information and estimate a cost-value function. The cost metric and its value estimation have been used to drive the learning process towards safety using penalties \cite{icra_safety, RCPO}, cumulative or instantaneous constraints \cite{Lagrangian, IPO}. However, the learned value functions have poor estimates and end up in local optima \cite{DeepPG, iclr2023}, limiting their efficacy in fostering safety. We argue that the cost's sparse nature is another key issue that hinders safety and sample efficiency. A sparse definition of the cost requires the agent to visit unsafe states to learn good estimates from the sparse feedback. These issues are pivotal in a safety context where we aim to minimize the number of visited unsafe states. 

\begin{figure}[b]
    \centering
    \vspace{-0.3cm}
    \includegraphics[width=0.6\linewidth]{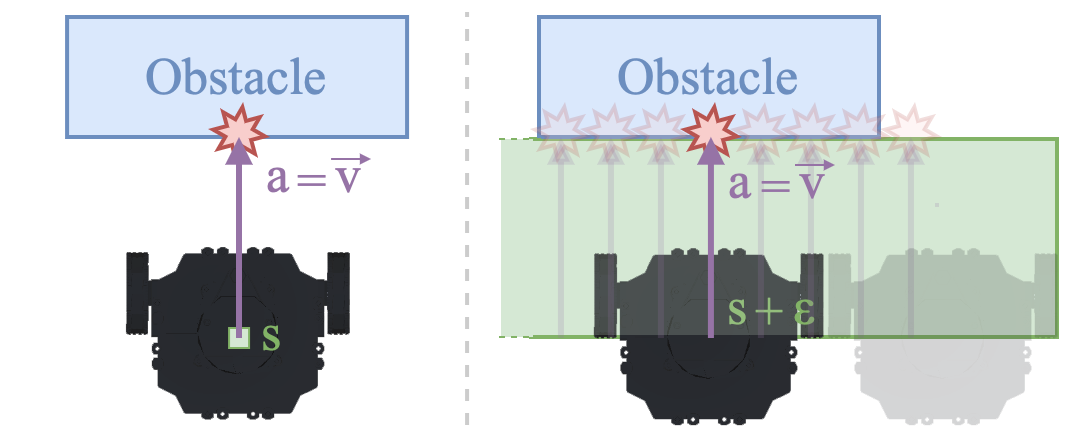}
    \vspace{-0.1cm}
    \caption{Indicator cost function (left). Unsafe interactions, caused by the same action, around the unsafe state (right).}
    \label{fig:visual_example}
    \vspace{-0.3cm}
\end{figure}

In this direction, we note that the indicator costs do not carry information about areas around $s$ where it is risky to perform the action that led to deeming $s$ unsafe. For example, consider a navigation scenario where a policy chooses the robot's velocity, given its position. In this context, prior works trigger a positive cost when colliding in a state $s$ (Figure \ref{fig:visual_example} on the left) and need to visit similar unsafe interactions around $s$ to learn the cost-value function that drives the learning process toward safety \cite{IPO, Lagrangian, Lyapunov}.
In contrast, it is possible to exploit high-level system specifications (e.g., robots' size and velocity) to define an area of size $\epsilon$ around $s$ where performing the action $a$ that led to collision would result in other unsafe interactions (Figure \ref{fig:visual_example} on the right). We can then compute a safety value based on the policy's decisions, avoiding visiting such an \textit{unsafe} area to learn a cost-value function. Hence, the idea is to replace indicator costs and the learning of cost-value functions by quantifying the states in the unsafe area where the policy chooses $a$ and use it as a penalty to discourage unsafe decisions \cite{RewardEngineering, RewardShaping}. Our hypothesis is that this procedure can significantly improve sample efficiency and reduce the number of visited hazardous states.

To this end, we propose an approximate \textit{violation} metric as a penalty that uses system specifications to quantify how safe policy decisions are around $s$. Following recent literature \cite{Verification, iros_aquatic, iros_ameya}, we encode whether a policy chooses specific actions (\textit{outputs}) in a subspace of the state space (\textit{inputs}) as input-output conditional statements, commonly referred to as task-level \textit{properties} (or state-action mappings). Formal Verification (FV) approaches for Deep Neural Networks (DNNs) have been used to formally check all the states where the policy violates such properties \cite{Neurify}. In particular, \cite{prove} introduced a formal violation metric by provably quantifying how well the policy respects the properties. Such a formal violation naturally addresses the limitation of indicator cost functions but has not been previously investigated to foster safety in DRL, as FV has two main issues. (i) It is a \textit{NP-Complete} problem \cite{reluplex} that makes it intractable to compute the formal violation metric at training time without a prohibitive overhead. (ii) The state-action mappings are hard-coded, which could be unfeasible in tasks with complex or unknown specifications. Against this background, we make the following contributions to the state-of-the-art:
\begin{itemize}
    \item We replace FV with a sample-based approach that approximates the violation metric with forward propagations of the agent's network on the sampled states. Such an approximation empirically shows a negligible error over the value computed with FV in a fraction of the computation time. 
    \item We generate an additional state-action mapping when performing an unsafe interaction during the training, using a fixed-size area around the visited unsafe state and the action that led to such interaction. 
    \item We show the advantages of using our approximate violation as a penalty in existing DRL algorithms and employ FV \cite{prove} on the trained policies to show that our approach allows learning safer behaviors (i.e., lower violations).
\end{itemize}
\noindent Our empirical evaluation considers a set of Unity \cite{unity} robotic mapless navigation tasks \cite{drl_navigation1, drl_navigation2, iros_marl}. In contrast to Safe DRL benchmarks (e.g., SafeMuJoCo \cite{IPO}), our scenarios allow transferring policies directly on the robot to foster the development of Safe DRL approaches in realistic applications. We also evaluate violation-based approaches in standard SafeMuJoCo tasks. In all scenarios, we compare with unconstrained DRL baselines augmented with a cost penalty \citep{DuelDDQN, PPO}, and constrained DRL \cite{Lagrangian}. Our evaluation shows that cost-based algorithms have higher costs and higher violations, confirming the lack of information provided by indicator cost functions. In contrast, the approximate violation-based penalty drastically reduces unsafe behaviors (i.e., lower cost and violation) while preserving good returns during and after training.

%% file: Sections/preliminaries.tex
\section{Preliminaries}
\label{sec:preliminaries}

A DRL problem is typically modeled as a Markov Decision Process (MDP), described by a tuple $<S,A,P,r,\gamma>$ where $S$ is the state space, $A$ is the action space, $P : S \times A \to S$ is the state transition function, $r : S \times A \to \mathbb{R}$ is a reward function, and $\gamma \in [0, 1)$ is the discount factor. In particular, given a policy $\pi \in \Pi := \{\pi(a|s) : s \in S, a \in A\}$, the agent aims to maximize the expected discounted return for each trajectory $\tau = (s_0, a_0,r_0, \cdots)$:
\begin{equation}
\max_{\pi \in \Pi} J_{r}^{\pi} := \mathbb{E}_{\tau \sim \pi} \left[\sum_{t=0}^\infty \gamma^t r(s_t, a_t)\right]
\end{equation}
\noindent A common approach to fostering safety is adding a penalty to the reward to learn how to avoid unsafe interactions characterized by lower payoffs \cite{Garcia2015}. 
Otherwise, MDPs can be extended to Constrained MDPs (CMDPs) to incorporate a set of constraints $\mathcal{C}$ defined on $C_{0,\dots,c}: S \times A \rightarrow \mathbb{R}$ cost functions (where $c$ is the number of constraints) and their thresholds $t_{0,\dots,c}$ \cite{ConstrainedMDP}. A $C_i$\textit{-return} is defined as $J_{C_i}^{\pi} := \mathbb{E}_{\tau \sim \pi} [\sum_{t=0}^\infty \gamma^t C_i(s_t, a_t)]$. Constraint-satisfying (feasible) policies $\Pi_\mathcal{C}$, and optimal policies $\pi^*$ are thus defined as:
\begin{equation}
    \Pi_\mathcal{C} := \{\pi \in \Pi : J_{C_i}^{\pi} \leq t_i,~\forall i \in [0,\dots,c]\}, \quad
    \pi^*= \max_{\pi \in \Pi_\mathcal{C}} J_{r}^{\pi}
\label{eq:cmdp}
\end{equation}
Constrained DRL algorithms aim at maximizing the expected return $J_r^{\pi}$ while maintaining costs under hard-coded thresholds $\mathbf{t}$:
\begin{equation}
\max_{\pi \in \Pi} J_{r}^{\pi} \quad\text{s.t.}\quad J_{C}^{\pi} \leq \mathbf{t}
\end{equation}
Such a constrained optimization problem is usually transformed into an equivalent unconstrained one using the Lagrangian method \cite{Nocedal, Lagrangian, LagrangianBehavior}, resembling a penalty-based approach. Recently, \citet{RCPO} also argued that constrained DRL has significant limitations as it requires a parametrization of the policy (i.e., it does not work with value-based DRL), a propagation of the constraint violation over trajectories and works with limited types of constraints (e.g., cumulative, instantaneous). For these reasons, the authors show the efficacy of integrating a penalty signal into the reward function. This motivates our choice of using penalty-based approaches to evaluate the benefits of incorporating the proposed approximate violation as a penalty to foster safety.
\subsection{Properties and Violation}
\label{sec:preliminaries_violation}
From FV literature \cite{Verification}, 
a property $P$ is hard-coded using task-level knowledge as a pre-condition $R$ and a post-condition $Q$ (i.e., $P := \langle R, Q \rangle$). In a DRL context, $R$ is the domain of the property (i.e., the area around $s$), and $Q$ is the co-domain (i.e., an action). Broadly speaking, given a DNN $\mathcal{N}$ with $y_{1,\dots, n}$ outputs, $R$ is defined with an interval $\epsilon_i$ for each input $i$ of $\mathcal{N}$, which we refer to as $\pmb{\epsilon}$-area, and $Q$ represents different desiderata on the output \cite{amir2022verifying}. 
For example, in a value-based setup where the policy selects the action with the highest value, the post-condition $Q$ of a property designed for safety models \textit{never select the action corresponding to the output} $y_k \in y_{1,\dots, n}$. Formally, $Q$ checks the following inequality:
\begin{equation}
    y_k < y_i~\forall i \in [1, n] - \{k\}
\label{eq:propcondition}
\end{equation}
that extends to continuous actions as shown in prior work \cite{Neurify, Verification}.

Recent works introduced a \textit{violation} metric to quantify the number of violations in the domain of the property using verification techniques \cite{prove, CountingProVe}. In more detail, the violation is defined as the ratio between the size of $R' \subseteq R$ where the post-condition is violated (i.e., the inequality does not hold and $y_k$ is selected) and $R$. Such a provable violation carries the task-level information of the properties and quantifies how often a property is violated. Hence, when using properties to model safety specifications, which we refer to as safety properties, the violation represents a locally-aware cost function.

%% file: Sections/methods.tex
\section{Methods}
\label{sec:method}

We aim to investigate the benefits of combining a reward and a violation-based penalty value into an MDP. Following prior penalty-based approaches \cite{Garcia2015, RCPO}, we maximize the following objective:
\begin{equation}
    \max_{\pi \in \Pi} J_{r,C}^{\pi} :=  \mathbb{E}_{\tau \sim \pi} \left[\sum_{t=0}^\infty \gamma^t r(s_t, a_t) - Z(\cdot)\right]
\label{eq:penaltyobj}
\end{equation}
\noindent where $Z(\cdot)$ is a generic penalty function. For example, a violation-penalty is $Z_\pi(s_{t}\pm\pmb{\epsilon})$, indicating that the violation depends on the policy decisions in a proximity $\pmb{\epsilon}$ of the state (i.e., the $\pmb{\epsilon}$-area, or $R$). Equation \ref{eq:penaltyobj} has two core benefits over other Safe DRL methods based on constraints: (i) penalty objectives potentially maintain the same optimal policy of the underlying MDP as they do not constrain exploration nor reduce the space of feasible policies as constrained approaches \cite{PolicyInvariance}.\footnote{This is not the case for navigation tasks as safe policies avoid obstacles.} (ii) Constrained DRL typically estimates an advantage function to propagate cost information, hindering their application with values that strictly depend on the current policy. Moreover, this requires visiting unsafe states to learn effective estimates for the sparse cost values. In contrast, penalty-based methods do not require a separate advantage estimate. 
In addition, various DRL algorithms, such as Proximal Policy Optimization (PPO) \cite{PPO}, provide significant empirical evidence of the benefits of using penalties instead of constraints \cite{RCPO}. 


\subsection{Approximate Violation}\label{sec:approx_vr}
A violation metric computed on a safety property quantifies the number of unsafe policy decisions over an area of the state space around the state $s$. Such a local component and the task-level knowledge inherited by the safety properties naturally address the indicator cost functions' lack of information.
However, the \textit{NP-Completeness} of FV \cite{reluplex} makes the provable violation computation intractable during the training due to the significant overhead of verification tools \cite{marabou}. Hence, we address the computational demands of FV by proposing a novel sample-based method to approximate the violation. 


Given a DNN $\mathcal{N}$ that parameterizes a policy, a list of properties $\mathbf{P} = \langle \mathbf{R}, \mathbf{Q} \rangle$, and the current state $s$, we aim at checking whether $y_k < y_i~\forall i \in [1, n] - \{k\}$ (where $\mathbf{y} := [y_{1, \dots, n}]$ are the outputs of $\mathcal{N}$ given $s$ as input). The general flow of our method is presented in Algorithm \ref{alg:approximate_violation}: first, we embed the condition (\ref{eq:propcondition}) in the network architecture by concatenating a new output layer that implements $y_i-y_k~\forall i \in [1, n]$ (line 1). We refer to this augmented network as $\mathcal{N}'$. If $s$ is contained in one or more pre-condition (i.e., it is deemed risky according to the properties), we consider such a subset of properties $\mathbf{P'} \subseteq \mathbf{P}$ to compute the approximate violation (line 2).
Hence, we randomly sample a set of states $I$ from the pre-conditions $\mathbf{R} \in \mathbf{P'}$ (line 3). Finally, after propagating $I$ through $\mathcal{N}'$, we enumerate the outputs $\mathcal{N}'(I) \leq 0$, which are the ones that do not satisfy the post-conditions (line 4). Finally, our approximate violation is the ratio between the number of such outputs over the total sampled points (line 5), which closely resembles how the formal violation is computed in \cite{prove}.

Similarly to the formal violation, our approximation can be interpreted as a locally-aware cost because it is computed using the information in a local region around a state. Moreover, our approximate violation: 
\begin{itemize}
    \item Includes safety information of areas of interest due to the state-action mappings (i.e., properties).
    \item Approximates how often a property violation might occur, having a similar role to Lagrangian multipliers but without requiring additional gradient steps or value estimators.
    \item It does not require additional environment interactions, drastically reducing the number of visited unsafe states. 
\end{itemize} 

\begin{algorithm}[t]
\caption{Computing the Approximate Violation}\label{alg:approximate_violation}
\quad\textbf{Given} $\mathcal{N}$ with outputs $y_{1, \dots, n}$, the current state $s$, properties $\langle \mathbf{R}, \mathbf{Q} \rangle$, and size $m$ of states to sample.
\begin{algorithmic}[1]
\State $\mathcal{N}' \leftarrow$ add a layer with $n$ outputs to $\mathcal{N}$ that implements (\ref{eq:propcondition})
\State $\mathbf{P'} \leftarrow \langle R, Q \rangle$ if $s \cap R \neq \varnothing,~\forall~\langle R, Q \rangle \in \langle \mathbf{R}, \mathbf{Q} \rangle$ \hfill $\rhd$ i.e., $s$ is unsafe
\State $\mathbf{I} \leftarrow $ Sample $m$ points from $p[R]~\forall p \in \mathbf{P'}$
\State $violation =$ Count $\max[\mathcal{N}'(I)] \leq 0$ $~\forall I \in \mathbf{I}$ \hfill $\rhd$ considering $y_k$ from $p[Q] ~\forall p \in \mathbf{P'}$
\State \textbf{return} $violation~/~(m~ |\mathbf{P'}|)$
\end{algorithmic}
\end{algorithm}
Finally, prior verification works only rely on hard-coded properties \cite{prove, crown, reluval}. Still, it is not uncommon to experience an unsafe state not included in the pre-conditions due to design issues, i.e., $s \cap R =\varnothing~\forall~R \in\mathbf{R}$. Hence, we generate an additional property upon experiencing an unsafe state $s$ using a fixed-size area around $s$ as $R$ and the performed action as $Q$. However, hard-coded properties are still crucial as there could be corner cases with more than one action to avoid.

\begin{figure}[b]
    \centering
    \includegraphics[width=1.0\linewidth]{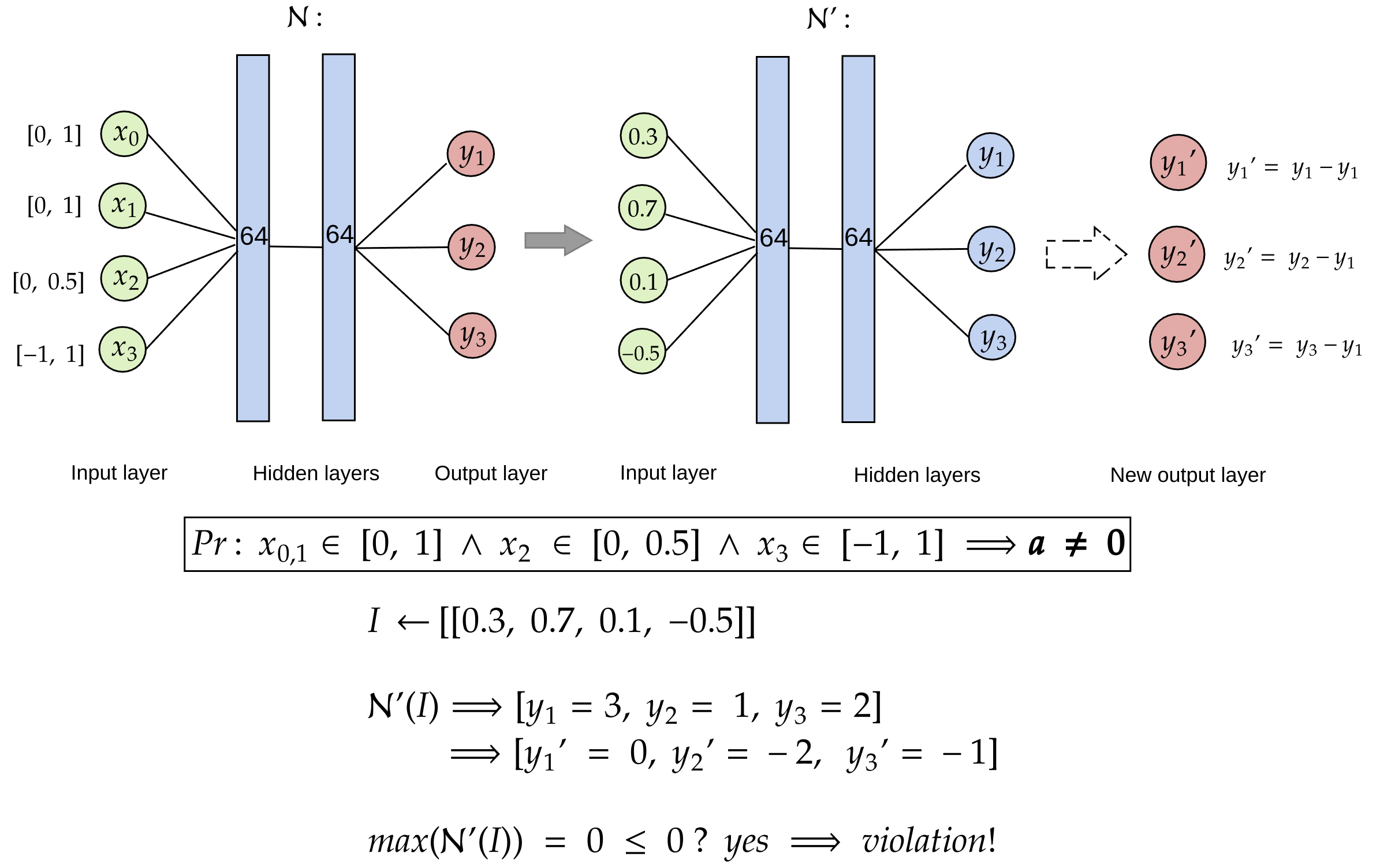}
    \caption{Example of computing the approximate violation.}
    \label{fig:example_violation}
\end{figure}

\subsubsection{Visual Example}\label{example_computation_violation}

We further detail the approximate violation computation using a visual example. In particular, Figure \ref{fig:example_violation} shows an illustrative example of a DNN and a property $Pr$ (following Section \ref{sec:design_prop} formalization). For the sake of simplicity, we show the process assuming to use only $m = 1$ sample from the property pre-condition. Following Algorithm \ref{alg:approximate_violation}, we show $\mathcal{N}'$ on the right of Figure \ref{fig:example_violation}, consisting of a new output that implements $y_i - y_k\; \forall i \in [1, n]$, where $y_k$ with $k = 1$ is the node that represents the action we are interested in avoiding. Considering the example in Figure \ref{fig:visual_example}, $x_{0, \dots, 3}$ are the current position, orientation, and distance from the goal for the robot. We want to avoid the action with index 0 (i.e., $y_1$), a forward movement at velocity $\Vec{v}$. We sample the $m = 1$ point from the property pre-condition, obtaining $i \in I$ that is then forward propagated through the network $\mathcal{N}'(I)$. Such propagation returns $y_1=3, y_2=1, y_3=2$ in the original output layer, and $y_1' = 0,\; y_2'=-2,\; y_3'=-1$ in the output of $\mathcal{N}'$. Finally, we enumerate the states $\in I$ where the maximum of $\mathcal{N}'(I)$ is less than or equal to 0, which means that $y_1$ will be chosen, leading to a violation. 

\subsection{Limitations}
\label{sec:limitations}
The violation requires hard-coded properties, which are challenging to design when considering agents with unknown dynamics. Our generated property does not consider scenarios where multiple actions are unsafe for the same state, so it remains unclear how to collect and refine properties during the training to model safe behaviors. Hence, as in FV literature \cite{Verification}, we assume having access to task-level knowledge to design the hard-coded properties. In safety-critical contexts, this assumption typically holds. Considering different input types (e.g., images) is conceptually feasible but would require further research and empirical evaluation. To this end, model-based DRL would allow using the model to design the unsafe area. Finally, it is unclear how to provide guarantees in model-free Safe DRL approaches, including our work, constrained DRL \cite{CPO, IPO, PCPO}, and several other approaches summarized in \citet{Garcia2015}. As discussed in \citet{DeepPG}, using DNNs for approximating policies and values makes the method diverge from the underlying theoretical framework. Nonetheless, we start addressing such key issues by employing existing FV approaches to check the trained policy decisions over the properties of interest.


%% file: Sections/experiments.tex
\section{Experiments}
\label{sec:experiments}

First, we introduce a set of Turtlebot3-based safety mapless navigation tasks to enable rapid testing of policies in realistic contexts. Our environments rely on Unity \cite{unity} as it allows rapid prototyping, Gym compatibility, and interface with the \href{www.ros.org}{Robotic Operating System (ROS)}. Mapless navigation is a well-known problem in model-free DRL \cite{drl_navigation1, drl_navigation2}, prior Safe DRL literature
\cite{SafetyGym, AAAI_Sos}, and multi-agent DRL \cite{icra_marl, iros_marl}. While standard navigation planners use a map of the environment and exhaustive information from the sensors, DRL setups consider more challenging conditions, such as not having a map and relying only on sparse local sensing.
We use a similar encoding to prior work \cite{drl_navigation1, drl_navigation2, drl_navigation3, marche2020discrete, curriculum}: 11 sparse laser scans with a limited range and two values for the target relative position (i.e., distance and heading) as observations. Discrete actions encode angular and linear velocities to reduce training times while maintaining good navigation skills \cite{marche2020discrete}.\footnote{Our environments also support continuous actions and different domain randomization of the tasks and physical properties through the Unity editor.} At step $t$, the agent receives a reward:
\begin{equation}
    r_t = \begin{cases}
    1 \quad \hfill \text{if goal reached} \\
    \Delta(d_{t-1}, d_{t}) - \beta \quad \hfill \text{otherwise}\\
    
    \end{cases}
\end{equation}
The agent thus obtains a dense reward given by the distance ($d$) difference ($\Delta$) from the goal in two consecutive steps, with a per-step $-\beta$ to incentive shorter paths. Each collision returns a positive cost signal that can be used to compute the desired penalty (e.g., violation), enabling a straightforward application to different penalty-based objectives (as in Equation \ref{eq:penaltyobj}) or constraints.

\begin{figure}[b]
    \centering
    \includegraphics[width=1.0\linewidth]{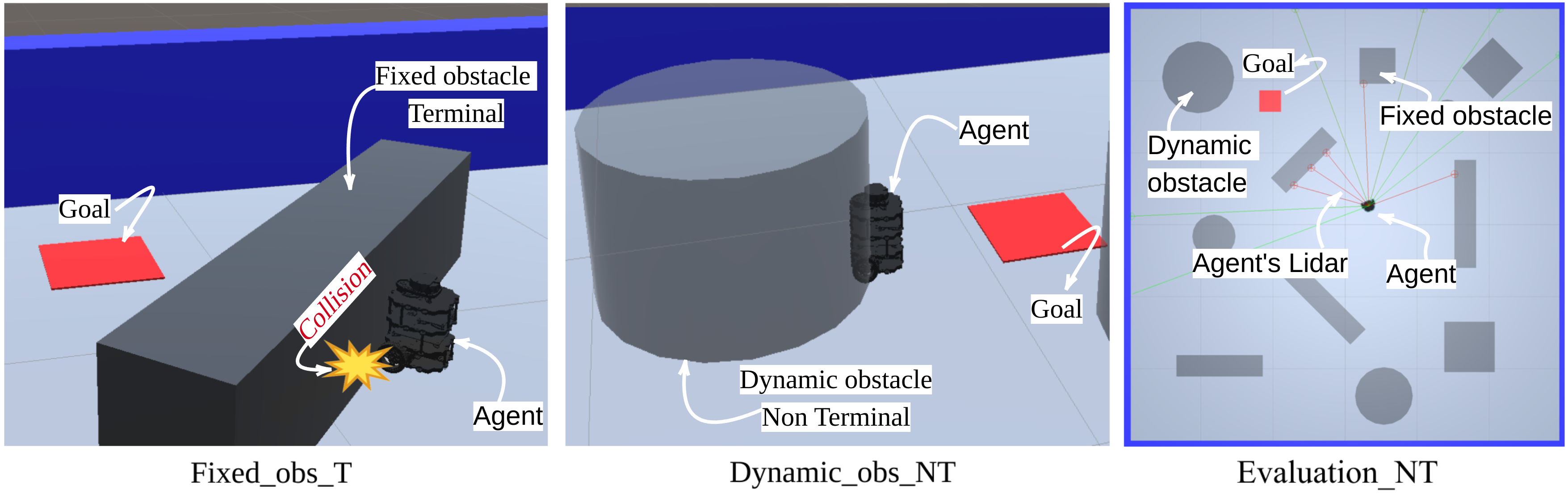}
    \caption{Fixed, Dynamic, Evaluation tasks with different obstacles. Terminal obstacles (T) reset the environment upon collision. Non-terminal ones (NT) allow the robot to cross them, experiencing more unsafe states. The evaluation environment has fixed and dynamic non-terminal obstacles.}
    \label{fig:exp_envs}
\end{figure}

We introduce four training and one testing environment with different obstacles, namely \textit{Fixed\_obs\_\{T, NT\}}, \textit{Dynamic\_obs\_\{T, NT\}}, and \textit{Evaluation\_NT} depicted in Figure \ref{fig:exp_envs}. Such a variety of conditions serve to provide different settings for evaluating Safe DRL algorithms in robotic navigation. The environments inherit several characteristics from known benchmarks such as SafetyGym \cite{SafetyGym} and differ from each other as the obstacles that can be \textit{Fixed} (parallelepiped-shaped static objects) or \textit{Dynamic} (cylindrical-shaped objects that move to random positions at a constant velocity). Moreover, obstacles can be \textit{Terminal} (T) if they end an episode upon collision, or \textit{Non-Terminal} (NT) if the agent can navigate through them.

\subsection{Environment Descriptions}
Our scenarios share a $4m\times4m$ size ($6m\times6m$ for the testing one), randomly generated obstacle-free goals, and a timeout at $500$ steps. A list of environments and their other main features follows:
\begin{itemize}
    \item \textit{Fixed\_obs\_T} has fixed terminal (T) obstacles. With terminal, we intend that the episode ends upon collision.
    \item \textit{Fixed\_obs\_NT} differs from the previous one for the non-terminal (NT) obstacles. The environment returns a signal upon each collision that can be used to model cost functions or other penalties. Non-terminal obstacles are visible to the lidars but non-tangible, i.e., the Turtlebot3 can pass through them. This class of obstacles represents the main challenge to designing safe DRL solutions, as the robot could get more positive rewards by crossing an obstacle at the expense of a higher number of unsafe behaviors.
    \item \textit{Dynamics\_obs\_T} has cylindrical-shaped dynamic terminal (T) obstacles. Such obstacles move toward random positions at a constant velocity, representing a harder challenge. The obstacles can travel on the robot's goal, so the agent must learn a wider variety of behaviors (e.g., react to an approaching obstacle, stand still to wait for the goal to clear).
    \item \textit{Dynamics\_obs\_NT}: differs from the previous one for the non-terminal (NT) obstacles.
    \item \textit{Evaluation\_NT}: we use this evaluation environment to test the generalization abilities of trained policies to new situations. This scenario is wider and contains both fixed and dynamic non-terminal obstacles of different shapes.
\end{itemize}

\begin{table*}[h]
    \centering
    \caption{Average \textit{violation} (\%), and computation time for properties $p_{\uparrow, \leftarrow, \rightarrow}$ calculated using ProVe\cite{prove}, and our approximation with 100, 1.000, and 10000 samples.}
    \label{tab:results:comparison}
    \begin{tabular}{lcccc}
      \toprule
      \bfseries Property & \bfseries ProVe & \bfseries Estimation 100 & \bfseries Estimation 1k & \bfseries Estimation 10k \\
      \midrule
      \textit{$p_{\uparrow}$} &  81.48 $\pm$ 1.2 & 81.0 $\pm$ 0.6 & 81.1 $\pm$ 0.8 & 81.17 $\pm$ 0.5 \\
       \textit{$p_{\leftarrow}$} & 73.9 $\pm$ 0.8  & 73.5 $\pm$ 0.2 & 73.63 $\pm$ 0.2 & 73.65 $\pm$ 0.3 \\
      \textit{$p_{\rightarrow}$} & 74.2 $\pm$ 0.3 & 73.6 $\pm$ 0.1 & 73.67 $\pm$ 0.1 &  73.68 $\pm$ 0.1 \\
      \midrule
      \textbf{Mean violation:} & \textbf{76.53} & 76.00 & 76.13 & \textbf{76.17} \\
      \midrule
      \textbf{Mean computation time:} & $\approx$\textbf{2m37s} & $\approx$\textbf{0.053s} & $\approx$ \textbf{0.056s} & $\approx$\textbf{0.060s} \\
      \bottomrule
    \end{tabular}
\end{table*}

\subsection{Properties for Mapless Navigation}\label{sec:design_prop}
Our properties shape rational, safe behaviors for mapless navigation and are used to compute the approximate violation. Moreover, we consider an online generated property described in Section \ref{sec:approx_vr}. A natural language description of the main hard-coded properties follows: 
\begin{itemize}
    \item $p_{\uparrow}$: \textit{There is an obstacle close in front $\Rightarrow$ Do not go forward}
    \item $p_{\leftarrow}$: \textit{There is an obstacle close to the left $\Rightarrow$ Do not turn left}
    \item $p_{\rightarrow}$: \textit{There is an obstacle close to the right $\Rightarrow$ Do not turn right}
\end{itemize}
We use the maximum agent velocity to determine the size of the area around the (unsafe) states of interest (i.e., the $\pmb{\epsilon}$-area). For example, a formal definition for $p_{\uparrow}$ is:
\begin{equation*}
\begin{split}
    p_{\uparrow}:\; &x_0,\dots,x_4 \in [0, 1] \wedge x_5\in [0, 0.05]\wedge x_6,\dots,x_{10} \in [0, 1] \wedge x_{11},\\
    &x_{12}\in [-1, 1] \implies a \neq 4
\end{split}
\end{equation*}

where $x_0, \dots, x_{10}$ are the 11 lidar values, $x_{11}, x_{12}$ is the relative position of the goal (i.e., distance and heading), and action $a=4$ corresponds to a forward movement. 
Crucially, each input $x_i$ potentially considers a different interval to model the area of interest. Hence, $p_{\uparrow}$ checks the policy's decisions when there is an obstacle close to the front (i.e., $x_5\in [0, 0.05]$) under any possible target position ($x_{11}, x_{12} \in [-1, 1]$).\footnote{We measured that a minimum normalized distance of $0.05$ is required for the robot to turn at max speed and avoid obstacles in front.} The approximate violation computed over these properties thus contains information about a specific unsafe situation under a general goal configuration (i.e., $x_{11}, x_{12} \in [-1, 1]$).

To assess how good our approximate violation is over the provable one, we compare it with the violation computed by a FV framework. In particular, the formal violation and our approximate one are computed using the above properties. They are averaged over the same ten models collected at random steps during the training. Table \ref{tab:results:comparison} shows the violation values for each property $p_{\uparrow, \leftarrow, \rightarrow}$ computed by ProVe \cite{prove}, and our approximation using 100, 1000, and 10000 samples. Our approximation shows an average $0.69\%$ error over the formal violation even when using only 100 samples. Such an error further decreases to $0.47\%$ by using 10000 samples. By exploiting parallelism and batch computation of modern Deep Learning Frameworks, the increase in computation time for the approximate violation with 100 or 10000 samples is comparable. Conversely, as discussed in Section 3.1, ProVe's average computation time is orders of magnitude higher with respect to our approach (i.e., 0.06 over 157 seconds). However, our approximate methodology does not formally guarantee the policy behaviors due to its sample-based nature. For this reason, the next section uses standard FV frameworks on the trained policies to show the provable violation results.

\begin{figure*}[t]
    \centering
    \includegraphics[width=0.8\linewidth]{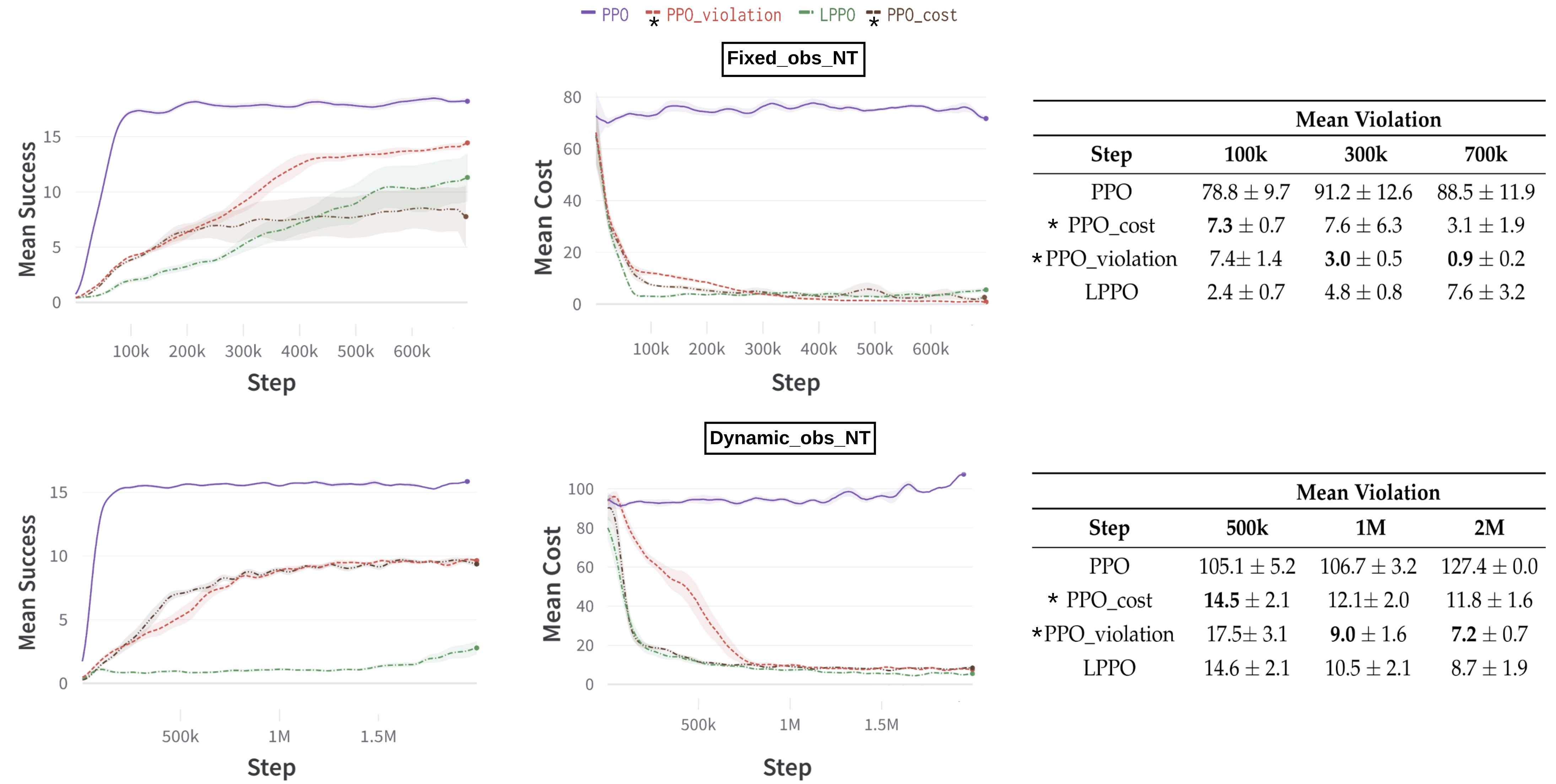}
    \caption{The two rows show average success, cost, and violation in the (NT) environments for PPO, the cost and violation penalty versions PPO\_\{cost, violation\}, and LPPO. * indicates penalty-based algorithms.}
    \label{fig:exp_results}
\end{figure*}

\subsection{Empirical Evaluation}
Our evaluation aims at showing the following: 

\begin{itemize}
    \item The benefits of integrating reward and safety specifications into a single MDP.
    \item The advantages of using our violation over indicator cost functions. To assess our claims, we plot the following values averaged over the last 1000 steps: (i) success (i.e., the number of goals reached), (ii) cost, (iii) and the violation at different stages of the training.
\end{itemize}

Data are collected on an i7-9700k and consider the mean and standard deviation of ten independent runs \cite{colas2019hitchhikers}. We consider the cost and violation penalty objective (\ref{eq:penaltyobj}) in a value-based (Dueling Double Deep Q-Network (DuelDDQN) \cite{DuelDDQN}) and policy-based (PPO \cite{PPO}) baselines, referring to the resultant algorithms as DuelDDQN\_\{cost, violation\} and PPO\_\{cost, violation\}. 
We compare with Lagrangian PPO (LPPO) \cite{Lagrangian} as it is a widely adopted Constrained DRL baseline and achieves state-of-the-art performance in similar SafetyGym navigation-based tasks.\footnote{For a fair comparison, we set the cost threshold of LPPO to the average cost obtained by PPO\_cost. The Appendices detail our hyper-parameters and are accessible at the following link: \url{shorturl.at/crSX6}} According to the literature, the value-based and policy-gradient baselines should achieve the highest rewards and costs (having no penalty information) \cite{SafetyGym}. Conversely, LPPO should show a significant trade-off between average cost and reward or fail at maintaining the cost threshold when set to low values \cite{IPO, SafetyGym, AAAI_Sos}. In contrast, we expect the penalty-based methods to achieve promising returns while significantly reducing the cost and the number of violations during the training. 

\textbf{Terminal Results.} Our results are shown in the Appendix. The information carried by the violation results in a significant performance advantage, maintaining similar or higher successes over non-violation-based approaches. Moreover, the policy-based algorithms show superior performance over the value-based implementations.

\begin{table*}[t]
\centering
    \caption{Average formal violation for the models at convergence in NT environment. Our violation-based penalty algorithm is the safest overall, as a lower violation value translates into fewer collisions.}
    \label{fig:suppl_verification}
    \includegraphics[width=0.8\linewidth]{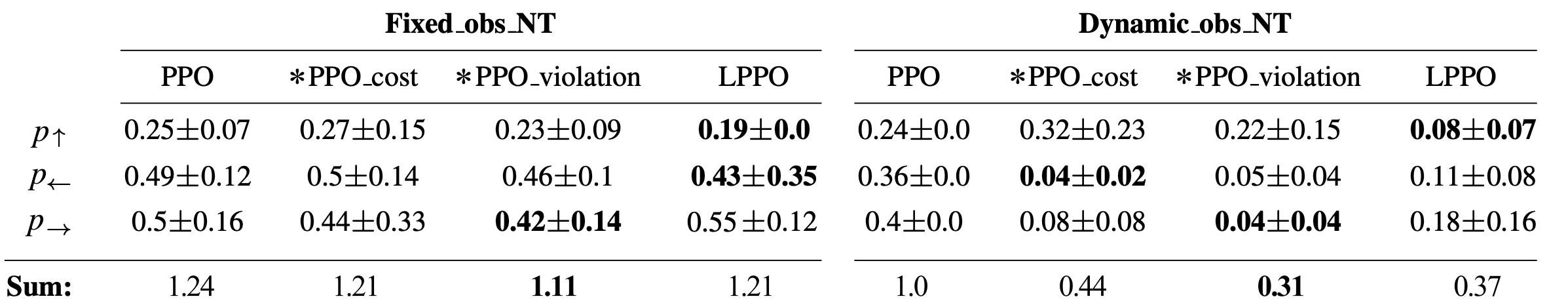}
\end{table*}

\textbf{Non-Terminal Results.} Given the higher performance of policy-based algorithms, the following experiments omit value-based ones. Figure \ref{fig:exp_results} show the results of (NT) tasks. As in the previous evaluation, PPO achieves a higher number of successes and a higher cost. In contrast, LPPO satisfies the constraint most of the time but achieves the lowest successes and does not learn effective navigation behaviors in \textit{Dynamic\_obs\_NT}, confirming the performance trade-off in complex scenarios \cite{SafetyGym, AAAI_Sos}. Moreover, PPO\_violation achieves better or comparable successes and cost values over PPO\_cost but significantly reduces the violations during the training. At convergence, PPO\_violation shows a $\approx2.2\%$ and $\approx4.6\%$ improvement over the cost counterparts, corresponding to 1320 and 2760 fewer unsafe policy decisions. In general, non-terminal tasks allow experiencing more unsafe situations in a trajectory, making the performance advantage in terms of the safety of violation-based algorithms more evident. 

In addition, we use FV \cite{prove} on the trained models of NT environments to provably guarantee the number of unsafe policy decisions over our safety properties. 
Table \ref{fig:suppl_verification} shows the average violations of the main properties for the models at the convergence of each training seed, which confirms that policies trained with PPO\_violation achieve lower violations, i.e., perform fewer collisions. The Appendix shows the same results for the T environments.

\begin{figure}[t]
    \centering
    \includegraphics[width=0.5\linewidth]{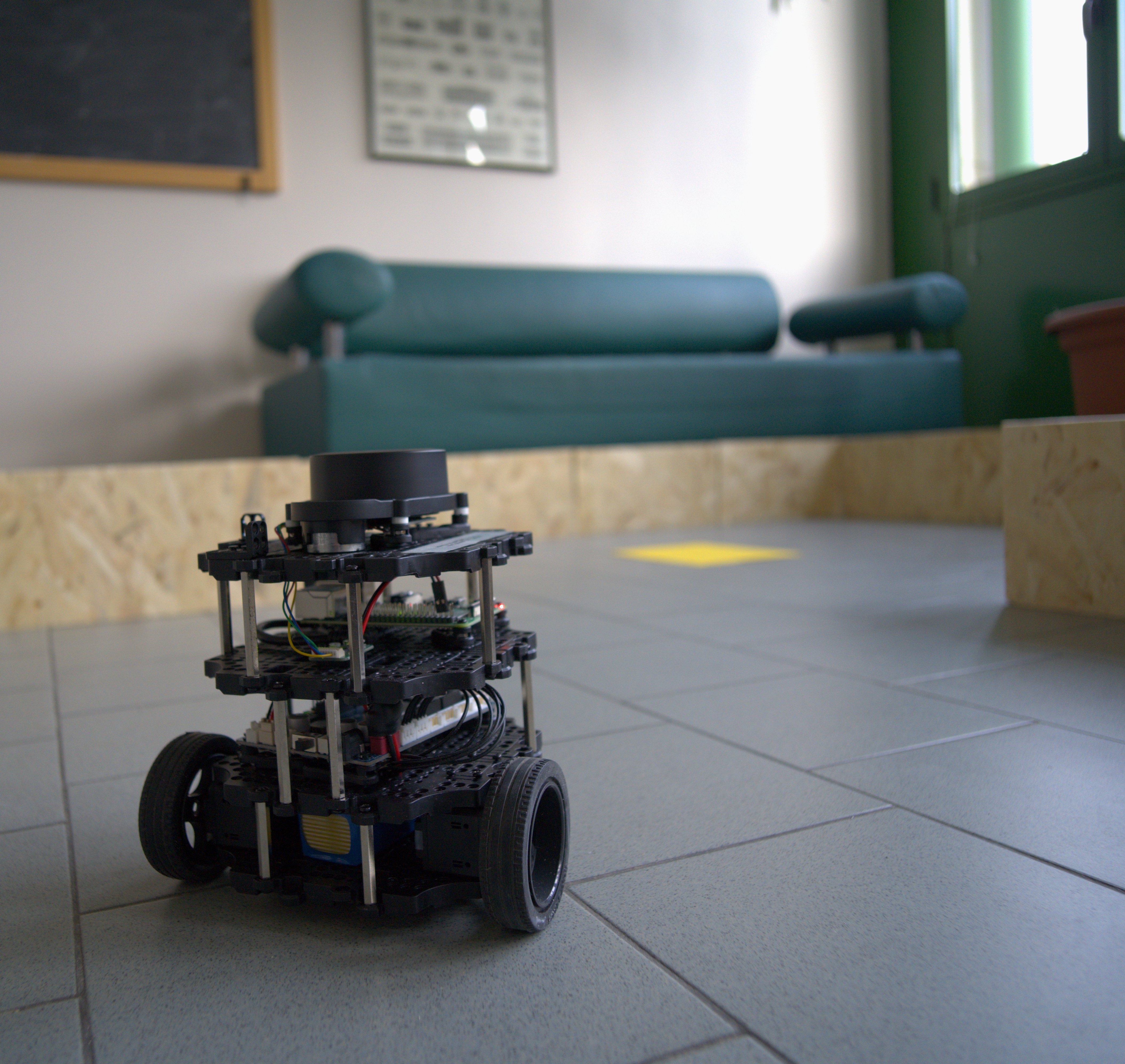}
    \captionof{figure}{Overview of real-world experiments.}
\label{fig:exp_real}
\end{figure}

\textbf{Evaluation Results.} Table \ref{tab:exp_eval} reports the average success, cost, and violation for the best model at the convergence of each training seed in the evaluation task of Figure \ref{fig:exp_envs}. The evaluation in a previously unseen scenario is used to test the generalization skills of the trained policies. In our experiments, the violation-based algorithm confirms superior navigation skills, achieving more success while being safer than LPPO and the cost counterpart. 
\begin{table}[t]
\centering
    \caption{Performance of the best-trained models in the testing environment \textit{Evaluation\_NT}.}
    \begin{tabular}{@{}lccc}
        \toprule
        & \multicolumn{1}{l}{Mean Success} & \multicolumn{1}{l}{Mean Cost} & \multicolumn{1}{l}{Mean Violation} \\ \midrule
        PPO & \textbf{10.0} $\pm$ 0.4 & 52.0 $\pm$ 4.8 & 63.7 $\pm$ 11.3 \\
        PPO\_cost & 6.4 $\pm$ 1.8 & 25.5 $\pm$ 8.3 & 29.2 $\pm$ 9.5 \\
        PPO\_violation & 7.7 $\pm$ 0.2 & \textbf{17.9} $\pm$ 1.2 & \textbf{18.8} $\pm$ 1.7 \\
        LPPO & 2.5 $\pm$ 0.3 & 18.2 $\pm$ 2.5  & 19.5 $\pm$ 0.9  \\
        \end{tabular}
    \label{tab:exp_eval}
\end{table}
\newline\textbf{Real-Robot Testing.} The core motivation for introducing our Safe DRL environments is the lack of benchmarks that allow testing the policies directly on real robots. In contrast, Unity environments enable the transfer of policies trained in simulation on ROS-enabled platforms. 
We report a video with the real-world experiments of the policies trained in our environments here: \url{shorturl.at/ijmFV}, while Figure \ref{fig:exp_real} shows our actual setup with the Turtlebot3.

\begin{figure*}[t]
    \centering
    \includegraphics[width=0.9\linewidth]{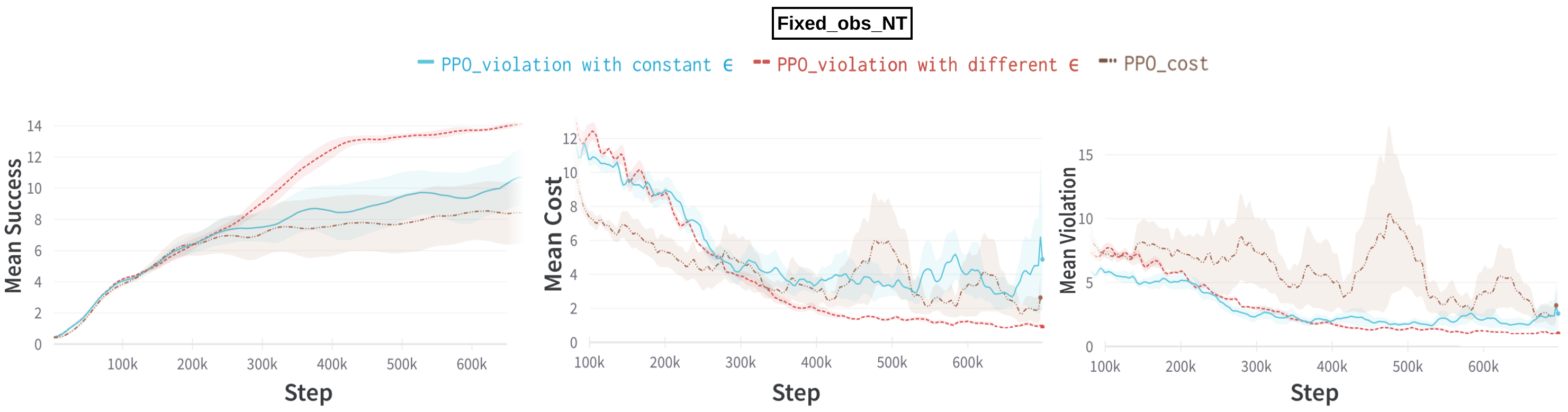}
    \caption{Average performance of penalty-based PPO using different pre-condition shapes: (i) fixed size (i.e., with a constant $\pmb\epsilon$), (ii) and one that considers the whole domain for the target's coordinates (i.e., different values for $\pmb\epsilon$. For example, $x_{11, 12} \in [-1, 1]$).}
    \label{fig:suppl_epsilonexp}
\end{figure*}

\begin{figure*}[t]
    \centering
    \includegraphics[width=0.9\linewidth]{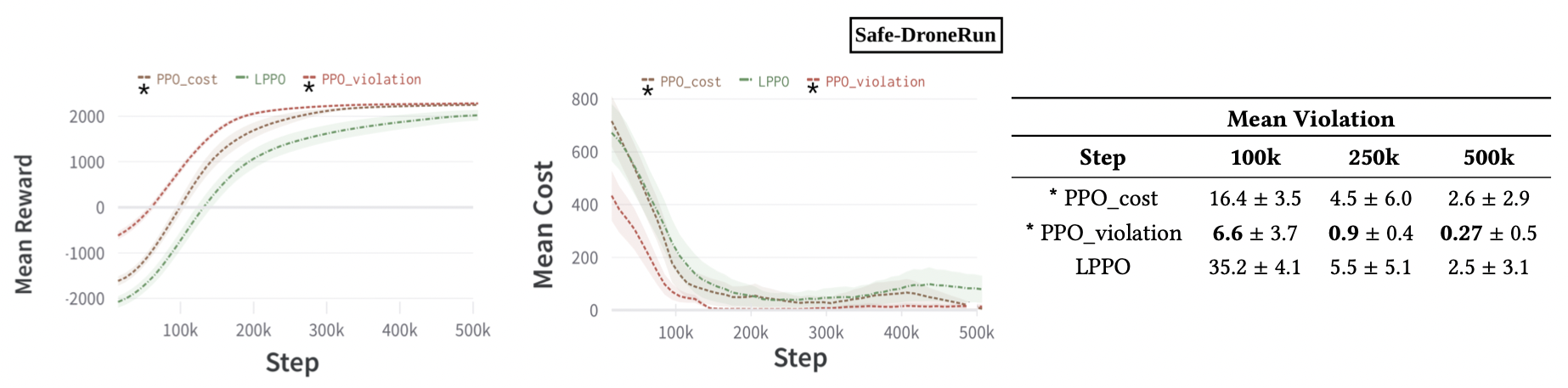}
    \caption{Average reward, cost, and violation in Safe-DroneRun for PPO, PPO\_\{cost, violation\}, LPPO. * indicates penalty-based algorithms.}
\label{fig:appendix_safetybullet}
\end{figure*}

\begin{figure}[t]
    \centering
    \includegraphics[width=.9\linewidth]{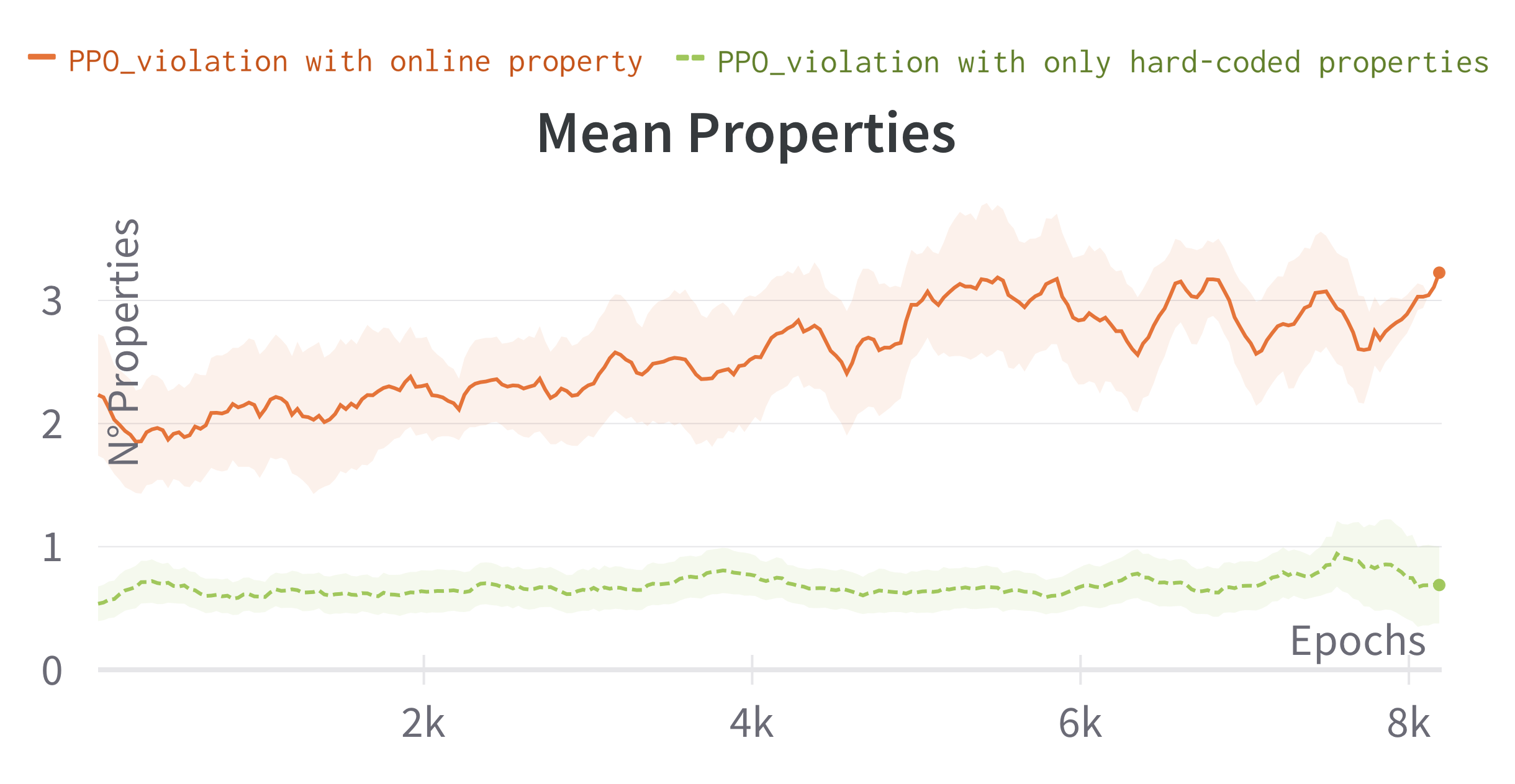}
    \caption{Mean size of $P'$ over the training for PPO\_violation with an online generated property (red), and with only the hard-coded properties (green).}
    \label{fig:suppl_nprop}
\end{figure}

\subsection{Additional Experiments}
We performed several additional experiments in the \textit{Fixed\_obs\_NT} task to highlight the impact of the different components presented in this paper using: (i) different sizes for the pre-conditions, (ii) the online property.\footnote{A regularization term can module the importance of the penalty over the training. We perform additional experiments with it in the Appendix.} (iii) We show the results of our violation-based penalty method in standard safe locomotion benchmarks to further confirm the performance improvement of the proposed approach.

\textbf{Different Sizes for Pre-Conditions.} Figure \ref{fig:suppl_epsilonexp} shows the results of two PPO\_violation versions: one that uses a constant $\pmb{\epsilon}$ for the size of the pre-conditions, and one that has different $\pmb{\epsilon}$ values for each input. The latter considers all the possible target positions as in previous experiments. As detailed in Section \ref{sec:design_prop}, using wider ranges for the inputs that shape environment configurations allows the violation to contain details about the unsafe behavior in general target initialization, resulting in higher performance. Crucially, the PPO\_violation with a constant $\pmb{\epsilon}$ also returns better performance over PPO\_cost, confirming the importance of having locally-aware information in the penalty value. This is particularly important as it may not be possible to shape detailed pre-conditions in setups where accurate task-level knowledge is lacking. 

\textbf{Online Property.}
As detailed in Section 3.1, we generate an additional property upon experiencing an unsafe state $s$ because it is not uncommon to experience an unsafe transition not included in the pre-conditions (due to the limitation of hard-coding properties). Figure \ref{fig:suppl_nprop} shows the size of the set $P'$ during the training for PPO\_violation under two implementations. The first adds an online-generated property, and the second uses only the hard-coded properties. Results for the latter show an average size of $P'$ $< 1$, confirming our hypothesis and the limitations of the properties' design of Section \ref{sec:limitations}. In contrast, the generated property implementation ensures having at least one input-output mapping for each unsafe state. This allows the violation to correctly shapes information about undesired situations, which biases the policy toward safer regions. Moreover, the growth in the size of $P'$ over the training indicates that the policy experiences unsafe states in rare corner cases captured by the intersection of multiple properties (i.e., some complex situations require not choosing more than one action to avoid collisions).

\textbf{Standard Safe DRL Tasks.}
We performed additional experiments in the standard Safe-DroneRun task to consider a different simulated robot, task, and safety specifications.
Our goal is to confirm further our framework's results and the benefits of using the violation as a penalty in a different domain known in the literature.\footnote{We refer to the original works for more details about the environments and the shaped cost functions \cite{CPO}, \url{github.com/SvenGronauer/Bullet-Safety-Gym}}
In more detail, we consider the same hyperparameters of our navigation experiments. Moreover, given the challenges of hand-writing properties for the drone, we rely only on the online generated property, considering a fixed $\epsilon$-area of size $0.05$ where we want to avoid a similar action that led to the unsafe state, up to decimal precision. 
The following results consider the average reward, cost, and violation collected over ten runs with different random seeds.

Figure \ref{fig:appendix_safetybullet} shows the results Safe-DroneRun. These results confirm the behavior of previous experiments (we omitted the PPO results to maintain the plot scale for better visualization), where LPPO struggles to keep the cost threshold set to 20 and results in lower performance compared to the penalty-based approaches. In contrast, the violation-based PPO maintains the best trade-off between reward and cost.

%% file: Sections/relatedwork.tex
\section{Related Work}
\citet{Garcia2015} presents an exhaustive taxonomy of the main families of approaches for Safe DRL, analyzing the pros and cons of each category. For example, model-based DRL approaches have been investigated in constrained and unconstrained settings \cite{ModelBasedSafe1, ModelBasedSafe2}. However, having access to or approximating a model is not always possible. Similarly, using barrier functions effectively fosters safety but requires an accurate system model \cite{BarrierSafe}.

In contrast, we focus on model-free learning. In this context, shielding approaches typically synthesize a shield (i.e., an automaton) to enforce safety specifications. However, this is usually unfeasible in complex setups due to the exponential growth in the size of the automaton. Hence, most DRL shielding approaches rely on simple grid-world domains \cite{Shielding, Shielding2}.
Although providing safety guarantees, it is unclear how to scale shielding approaches in complex, realistic applications. In contrast, constrained DRL has been used as a natural way to address Safe DRL \cite{CPO, PCPO, FOCOPS, Lyapunov2, Lagrangian}. In detail, CPO \cite{CPO} is characterized by strongly constrained satisfaction at the expense of possibly infeasible updates that requires demanding second-order recovery steps. Similarly, PCPO \cite{PCPO} uses second-order derivatives and has mixed improvements over CPO \cite{FOCOPS}.
Moreover, Lyapunov-based algorithms \cite{Lyapunov2} combine a projection step with action-layer interventions to address safety. However, the cardinality of Lyapunov constraints equals the number of states, resulting in a significant implementation effort. Despite the variety of constrained literature, we compare with Lagrangian methods \cite{Lagrangian} as they reduce the complexity of prior approaches and show promising constraints satisfaction. However, constrained DRL has several drawbacks. For example, incorrect threshold tuning leads to algorithms being too permissive or, conversely, too restrictive \cite{Garcia2015}. Moreover, the guarantees of such approaches rely on strong assumptions that can not be satisfied in DRL, such as having an optimal policy. Constrained DRL is thus not devoid of short-term fatal behaviors as it can fail at satisfying the constraints \cite{AAAI_Sos}. Moreover, constraints naturally limit exploration, causing getting stuck in local optima or failing to learn desired behaviors properly \cite{DiversityExpl, gecco}.

We note that our Equation \ref{eq:penaltyobj} falls under the category of reward engineering, which has been proved effective by several works \cite{Garcia2015, Reward, RewardEngineering, RewardShaping}. For example, IPO \cite{IPO} uses a penalty function based on constraints, giving a zero penalty when constraints are satisfied and a negative infinity upon violation. However, tuning both the barrier parameter and the constraint threshold has sub-optimal solutions over Lagrangian methods \cite{AAAI_Sos}. 
Finally, Statistical Verification (SV) has been recently employed on learning systems \cite{StatisticalVerification} to deal with the computational demands of FV. In these approaches, desired specifications are defined as Signal Temporal Logic \cite{STL}, which closely resembles the properties used by FV literature. However, a distinctive feature of our method is the use of the violation value, which can not be directly computed by using SV.


%% file: Sections/discussion.tex
\section{Discussion}
We present an unconstrained DRL framework that leverages local violations of input-output conditions to foster safety. We discussed the limitations of using cost functions as in Safe DRL \cite{Garcia2015} presenting: (i) a sample-based approach to approximate a violation metric and use it as a penalty in DRL algorithms. Such a violation introduces task-level safety specifications into the optimization, addressing the cost's lack of information. (ii) The influence of generating properties to cope with the limitations of hard-coded conditions. (iii) We argued the importance of developing real-world environments for broader applications of DRL in practical scenarios. To this end, we presented an initial suite of robotic navigation tasks that allow rapid testing on ROS-based robots. 

This work paves the way for several research directions, including extending our task suite to create a general safe DRL benchmark. Such extension is possible due to the rapid prototyping benefits of Unity \cite{unity}. Moreover, studying the effects of different time horizons would be interesting to separate the importance given to rewards and safety values. It would also be interesting to design a shield to avoid unsafe behaviors at deployment, leveraging the information from FV. Finally, our insights on the violation could be used to model desired behaviors in single and multi-agent applications to improve performance and sample efficiency.